\documentclass[conference]{IEEEtran}

\usepackage{amsmath}
\usepackage{algorithm}
\usepackage{algpseudocode}
\usepackage{subcaption}
\usepackage{array}
\usepackage{tabularx}  
\usepackage{float}
\usepackage{booktabs}
\usepackage{multirow}
\usepackage{graphicx}
\usepackage{cite}
\usepackage{array}
\usepackage{url}

\begin{document}

\title{Robust Lightweight Deep Learning Models for Oral Cancer Screening }

 \author{
 \IEEEauthorblockN{Siddhant Bharadwaj\IEEEauthorrefmark{1}\IEEEauthorrefmark{2}}
 \IEEEauthorblockA{\textit{Indian Institute of Science} \\
 siddhant0701@gmail.com}
 \and
 \IEEEauthorblockN{Aakash Shedsale\IEEEauthorrefmark{1}}
 \IEEEauthorblockA{\textit{Indian Institute of Science} \\
 aakashsunil@iisc.ac.in}
 \and
 \IEEEauthorblockN{Tejashree Subramanya\IEEEauthorrefmark{1}}
 \IEEEauthorblockA{\textit{Indian Institute of Science} \\
 stejashree@iisc.ac.in}

 \and
 \IEEEauthorblockN{Mohd. Azfar}
 \IEEEauthorblockA{\textit{Indian Institute of Science} \\
 azfarmohd@alum.iisc.ac.in}
 \and
 \IEEEauthorblockN{Praveen Birur}
 \IEEEauthorblockA{\textit{KLES’ Institute of Dental Sciences} \\
 praveenbirur@klesids.edu.in}
 \and
 \IEEEauthorblockN{Debnath Pal}
 \IEEEauthorblockA{\textit{Indian Institue of Science} \\
 dpal@iisc.ac.in}
 \and
 \IEEEauthorblockN{Shankararama Sharma}
 \IEEEauthorblockA{\textit{Indian Institue of Science} \\
 shankararama@iisc.ac.in}
 \and
 \IEEEauthorblockN{Anupama Shetty}
 \IEEEauthorblockA{\textit{Biocon Foundation} \\
 anupama.shetty101@biocon.com}
 \and
 \IEEEauthorblockN{Rajesh Sundaresan}
 \IEEEauthorblockA{\textit{Indian Institute of Science} \\
 rajeshs@iisc.ac.in}
 \\[0.5em]
 \IEEEauthorrefmark{1}Equal contribution \quad \IEEEauthorrefmark{2}Corresponding author

 }

\maketitle

\begin{abstract}
Oral cancer is a leading cause of mortality in low-to-middle-income countries, where a shortage of specialists delays diagnosis. While point-of-care screening via smartphones offers a scalable solution, developing robust AI for resource-constrained settings poses significant challenges, including class imbalance in training data, variable data quality, and computational constraints on edge devices. In this paper, we present the optimisation of lightweight deep learning models for smartphone-based oral cancer screening. Using a diverse, multi-centre retrospective dataset of approximately 30,000 images acquired over a decade, we systematically evaluate state-of-the-art convolutional, transformer, and hybrid architectures. Through rigorous pipeline ablation, we demonstrate that directly optimising hybrid architectures for the edge strictly outperforms computationally heavy paradigms, such as large models or knowledge distillation. Furthermore, interpretability analysis and simulated noise-stress tests revealed that the system anchors on clinical features and remains robust to unstructured sensor noise, despite vulnerabilities to impulse bit errors. In the held-out test set, our optimised MobileViTv2 models achieved an average sensitivity of 83.2 $\pm$ 1.5\% and an average specificity of 86.0 $\pm$ 0.8\%, with the best model exhibiting 87.4\% sensitivity, 86.5\% specificity, and a critical negative predictive value of 97.2\% with reference to specialist labels. These results confirm that with targeted architectural selection and streamlined optimisation, interpretable and robust lightweight AI models exhibit high potential for edge deployment to enable automated triage in primary care settings.
\end{abstract}


\section{Introduction}

Oral cancer is a major public health challenge and a leading cause of mortality in low-to-middle-income countries. In India alone, over 140,000 new cases are diagnosed annually, with the majority detected at advanced stages due to limited access to specialised care \cite{ferlay2024, mignogna2001}. Although visual examination of the oral cavity is an established method to detect oral potentially malignant disorders (OPMDs), asymptomatic but clinically evident oral lesions precede the onset of oral cancer \cite{rajaraman2015recommendations, ranganathan2019oral}, field studies indicate that primary healthcare workers achieve variable sensitivities, sometimes as low as 43.6\% compared to specialists \cite{birur2022field}. This diagnostic gap underscores the urgent need for automated, point-of-care, adjunctive tools.

To address this, recent studies have demonstrated the feasibility of using smartphone-based imaging combined with deep learning to classify oral lesions at the point of care \cite{birur2022field, lin2021automatic, song2024classification, talwar2023ai}. However, these works have largely relied on small, curated datasets and lack systematic evaluations of how models perform under real-world constraints. In real-world primary care settings, data is heavily affected by class imbalance (normal cases far outnumbering suspicious ones) and physical acquisition artefacts, such as blur, poor lighting, and varied camera angles. Currently, there is a lack of frameworks that evaluate the robustness and interpretability of lightweight edge-AI architectures against these specific, real-world challenges.

In this study, we bridge this gap by systematically developing and optimising robust, lightweight deep learning systems for oral cancer screening. We use a diverse, multi-centre retrospective dataset of approximately 30,000 smartphone-captured images collected over a decade. We highlight the technical challenges inherent in the dataset, and demonstrate how to use it effectively to build a light-weight model.

The main contributions of this paper are:
\begin{itemize}
    \item \textbf{Edge-Optimised Architecture Identification:} We conduct a rigorous evaluation of state-of-the-art convolutional, transformer, and hybrid architectures, identifying MobileViTv2 as the optimal inference engine. It successfully balances the global contextual awareness of transformers with the local feature extraction of convolutional layers.
    \item \textbf{Streamlined Pipeline Ablation:} We demonstrate through rigorous ablation studies that directly optimising lightweight models on ground-truth labels outperforms compute-heavy, large model paradigms. We explicitly establish that knowledge distillation introduces overhead without significant improvement in performance.
    \item \textbf{Interpretability Validation:} We validate the system's diagnostic integrity through GradCAM++ and multi-scale attention rollout, confirming that the network largely anchors its predictions on annotated clinical features rather than spurious background or hardware artefacts.
    \item \textbf{Hardware Stress-Testing and Clinical Efficacy:} We stress-test the operational bounds of the system against simulated hardware-induced signal degradations (e.g., low-light sensor noise, impulse bit-errors). Ultimately, we present a robust framework that approaches specialist-level triage (87.4\% sensitivity, 97.2\% Negative Predictive Value) with on-device inference capabilities.
\end{itemize}


\section{Materials and Methods}

\subsection{Retrospective Data}
A retrospective dataset was used to develop, tune, and evaluate the AI models. To ensure a robust evaluation of model generalisability, the cleaned dataset was partitioned into a training set for model optimisation, a validation set for checkpointing, and a strictly held-out test set for final performance benchmarking.

\subsubsection{Data Collection}

The dataset comprised images sourced from oral cancer outreach programmes conducted by the Biocon Foundation across the northern, north-eastern, and southern states of India between 2011 and 2023. These programmes primarily targeted rural populations. Crucially, from a systems perspective, the images were captured using a heterogeneous mix of consumer-grade smartphones (including HTC, Motorola, Xiaomi, and Samsung devices) operated by frontline health workers. The camera resolutions ranged from 5 to 16 megapixels. This hardware diversity inherently introduced real-world acquisition variations such as differing sensor noise profiles, colour calibrations, resolutions and focal lengths, providing an ideal testbed for evaluating the robustness of edge-AI models under diverse acquisition conditions.

\subsubsection{Data Description}
The raw dataset of 31,601 images was annotated by two oral medicine specialists. As part of a rigorous quality-control pipeline, statistically identical duplicates (n=126) and non-diagnostic images (e.g., extreme blur, out-of-focus, or insufficient mouth opening) were removed prior to model training. The final cleaned dataset comprised 29,574 images, each with an associated provisional clinical diagnosis that was subsequently mapped into a binary classification schema: \textit{Suspicious} and \textit{Non-suspicious} (Table \ref{tab:provisional-diagnosis}). The \textit{Suspicious} category includes oral lesions requiring referral (from potentially malignant disorders to carcinomas). Conversely, the \textit{Non-suspicious} category includes normal anatomy, normal variations, and benign presentations. This aggregated categorisation introduces significant intra-class variance within both the positive and negative classes, presenting a non-trivial feature-extraction challenge for the deep learning (DL) models. 

\begin{table}[htbp]
\centering
\caption{Provisional diagnosis label categorisation}
\label{tab:provisional-diagnosis}
\begin{tabular}{|l|r|l|}
\hline
\textbf{Provisional Diagnosis} & \textbf{Count} & \textbf{Category} \\ \hline
Normal/Normal variations & 22438 & \multirow{3}{*}{Non-suspicious} \\
Benign & 870 & \\
Other & 1379 & \\ \hline
Homogenous Leukoplakia & 1828 &\multirow{7}{*}{Suspicious} \\
Oral Lichen Planus & 578 & \\
Tobacco Pouch Keratosis & 1898 & \\
Oral Submucousal Fibrosis & 28 & \\
Non-homogeneous Leukoplakia & 395 & \\
Verrucous Leukoplakia & 16 & \\
Malignant & 144 & \\ \hline
\end{tabular}

\vspace{6pt}

\caption{Datasplit demonstrating real-world class imbalance}
\label{datatab}
\begin{tabular}{|l|r|r|r|}
\hline
\textbf{Split} & \textbf{Suspicious} & \textbf{Non-suspicious} & \textbf{Total} \\
\hline
Train      & 2932 & 14812 & 17744 \\
Validation &   978 &  4937 &  5915 \\
Test       &   977 &  4938 &  5915 \\
\hline
\textbf{Total} & \textbf{4887} & \textbf{24687} & \textbf{29574} \\
\hline
\end{tabular}
\end{table}

\subsubsection{Datasplit and Class Imbalance}
The curated dataset was divided into a 60-20-20 split for training, validation, and testing (Table \ref{datatab}). Representative of real-world screening scenarios, the dataset exhibits a severe class imbalance (approximately 1:5 ratio of suspicious to non-suspicious images), necessitating targeted optimisation strategies during training.

\subsection{Lightweight AI Architectures}
To ensure that the resulting models are computationally viable for edge deployment on budget smartphones, we restricted our architectural search to lightweight models with parameter counts under 7.5 million. We evaluated six distinct architectures (Table \ref{tab:models}), spanning Convolutional Neural Networks (EfficientNetV2-B0, MobileNetV3-Large, NASNet-Mobile), pure Vision Transformers (DeiT-Ti), and hybrid CNN-Transformer models (MobileViTv2, EdgeNeXt-S) \cite{mehtaseparable, maaz2022edgenext, touvron2021training,  howard2019searching, tan2021efficientnetv2, zoph2018learning}. This selection allows us to systematically compare the inductive biases of CNNs against the global context modelling of Transformers for use in resource-constrained environments.

\begin{table}[htbp]
\caption{Summary of evaluated lightweight AI models}
\label{tab:models}
\centering
\resizebox{\linewidth}{!}{%
\begin{tabular}{|m{2cm}|>{\centering\arraybackslash}m{1.3cm}|>{\centering\arraybackslash}m{1cm}|>{\centering\arraybackslash}m{1cm}|>{\centering\arraybackslash}m{3cm}|}
\hline
\textbf{Model} & \textbf{Type} & \textbf{Params} & \textbf{Input size} & \textbf{Key design} \\
\hline
MobileViTv2-1.0 & Hybrid & 4.9M & 256x256 & Separable self-attention for efficient contextual learning \\ \hline
EdgeNeXt-S & Hybrid & 5.6M & 256x256 & Adaptive convolution and split depth-wise transpose attention (SDTA) \\ \hline
DeiT-Ti & Transformer & 5.0M & 224x224 & Distillation-based training with teacher-student framework \\ \hline
NASNet-Mobile & CNN & 5.3M & 512x512 & Neural architecture search (NAS)-designed with depthwise separable convolutions \\ \hline
MobileNetV3-L & CNN & 5.4M & 512x512 & Inverted residuals, squeeze-and-excite modules, h-swish activation \\ \hline
EfficientNetV2 & CNN & 7.1M & 512x512 & Fused-MBConv blocks with squeeze-and-excite for efficiency \\
\hline
\end{tabular}%
}
\end{table}

\subsection{Experimental Setup}
\label{expts}

\subsubsection{Training Infrastructure and Hyperparameters}
All models and ablation studies were trained on an on-premise compute node with 3 NVIDIA RTX 3090 GPUs using PyTorch 2.3 framework. Models were initialised with ImageNet pre-trained weights \cite{deng2009imagenet}. Input images were normalised using ImageNet's channel-wise mean and standard deviation and resized via bicubic interpolation with corner alignment. Training was performed for 50 epochs with a batch size of 32, and the checkpoint with the best validation performance was selected for test-set evaluation. Each experimental configuration was repeated 10 times with different fixed random seeds, and the mean and variance of the observed performance on the test set are reported.

\subsubsection{Data Augmentations for Real-World Robustness}
To simulate the high variability of smartphone-captured images and prevent over-parameterisation, conventional geometric and photometric augmentations were applied, including zoom-in, colour jitter, and horizontal/vertical flips with random rotations between -10$^\circ$ to 10$^\circ$. 

Furthermore, to encourage lightweight models to focus strictly on the oral cavity rather than extraneous background artefacts, we explored whether bounding boxes generated by the Segment Anything Model (SAM) \cite{kirillov2023segment} can improve performance when used as an augmentation. Because accurate bounding boxes were unavailable for all images, SAM was deployed in the ``everything'' mode to generate coarse masks that removed non-diagnostic regions. It is important to note that SAM was utilised strictly during the offline training pipeline and is not part of the inference workflow.

\subsubsection{Image Quality Reweighting}
Smartphone images captured in primary care settings often suffer from poor lighting and inconsistent focus. To investigate whether integrating perceptual image quality into the loss function could emphasise learning from high-quality images and penalise learning from low-quality ones, we explored reweighting strategies using two no-reference image quality assessment (IQA) metrics: the Naturalness Image Quality Evaluator (NIQE) and CLIP-IQA \cite{mittal2012making, wang2023exploring}.

NIQE evaluates deviations from natural scene statistics (NSS) and produces a scalar score where lower values indicate higher quality. During training, the inverse of the NIQE score was integrated into the loss function to dynamically down-weight low-quality images. Conversely, CLIP-IQA assesses perceptual quality based on semantic consistency with natural image priors. By using CLIP-IQA scores as weighting factors, we tested whether high-level feature understanding could guide the model to prioritise images that are perceptually closer to high-quality references.

\subsubsection{Handling Systemic Class Imbalance}
The retrospective dataset exhibits a systemic class imbalance typical of screening environments, with one suspicious image for every five non-suspicious images in the training set (2,932 vs. 14,812). To prevent the models from collapsing into majority-class prediction, we systematically evaluated five mitigation strategies:

\begin{itemize}
    \item \textbf{Random Undersampling:} The majority (non-suspicious) class was randomly undersampled to match the minority class count.
    
    \item \textbf{Random Oversampling:} The minority (suspicious) class was stochastically oversampled.
    
    \item \textbf{Weighted Cross-Entropy (WCE):} WCE assigns a higher penalty to
    minority-class misclassifications by weighting the loss inversely to class
    frequency~\cite{aurelio2019learning}:
        \begin{equation}
            \label{eq:WCELoss}
            WCE(p;\theta) = - w_0 (1-p_t)\log(1-p(\theta)) - w_1 p_t \log(p(\theta))
        \end{equation}
    where $p_t \in \{0,1\}$ denotes the true label (1 for suspicious), $p(\theta)$ is
    the predicted probability, and $w_0, w_1$ are the class weights. Following the
    inverse-frequency scheme $w_c = N/(2N_c)$ on the training split
    (Table~\ref{datatab}), these are set to $w_0 = 0.60$ (non-suspicious) and
    $w_1 = 3.03$ (suspicious).

    \item \textbf{Focal Loss:} To prevent easy-to-classify background examples from dominating the gradient, Focal loss down-weights well-classified examples and focuses on hard examples \cite{ross2017focal}:
    \begin{equation}
        \label{eq:FocalLoss}
        \begin{split}
        FL(p;\theta) = & - (1-p_t)(p(\theta))^\gamma \log(1-p(\theta)) \\
                       & - (p_t)(1-p(\theta))^\gamma \log(p(\theta))
        \end{split}
    \end{equation}
    where $\gamma$, the focusing parameter, is set to $\gamma=2$.
    
    \item \textbf{Balanced Mix-up:} A hybrid regularisation and sampling technique that simultaneously performs instance-based and class-balanced sampling \cite{galdran2021balanced}, mixing them to create a smoother decision boundary:
   \begin{equation}
    \label{eq:BalancedMixup}
    \begin{aligned}
    \hat{x} &= \lambda x_I + (1- \lambda) x_C \\
    \hat{y} &= \lambda y_I + (1-\lambda) y_C
    \end{aligned}
    \end{equation}
    where $(x_I, y_I)$ are standard instance-based samples, $(x_C, y_C)$ are class-balanced samples, with $x$ denoting images and $y$ denoting the ground truth labels, and the mixing coefficient $\lambda \sim \text{Beta}(\alpha,1)$, where $\alpha$ is a hyperparameter, which was set to 0.1, 0.2 and 0.3 in our experimentations.
\end{itemize}

\section{Experimental Results}
\label{results}
The results of all the experiments detailed in Section \ref{expts} are explained in this section. The performance metric `Best Pair', mentioned in Tables \ref{augmentation_performance_all_models}, \ref{reweighting_results}, \ref{tab: class_imbalance_results}, \ref{crossval}, \ref{tab:combined_kd}, corresponds to the sensitivity and specificity of the top-performing model (in terms of geometric mean (GM)) among the 10 fold-wise checkpointed models.

\subsection{Data Augmentation and Pipeline Efficiency}
\label{augmentations}
\begin{table*}[htbp]
\caption{Performance of models across various augmentation strategies}
\centering
\label{augmentation_performance_all_models}
\renewcommand{\arraystretch}{1.25}
\begin{tabular}{|l|l|c|c|c|c|c|c|}
\hline
\textbf{Augmentation} & \textbf{Metric} & \textbf{MobileViTv2} & \textbf{EdgeNeXt-S} & \textbf{DeiT-Ti} & \textbf{MobileNetV3-L} & \textbf{EffNetV2-B0} & \textbf{NASNet-Mobile} \\ \hline
\multirow{4}{*}{Baseline} & Sensitivity & 75.48$\pm$3.69 & 72.01$\pm$3.48 & 69.39$\pm$4.25 & 59.48$\pm$12.46 & 61.03$\pm$8.50 & 72.11$\pm$3.85 \\
 & Specificity & 90.45$\pm$3.03 & 90.42$\pm$2.61 & 91.52$\pm$1.68 & 93.99$\pm$2.61 & 94.91$\pm$2.50 & 90.95$\pm$2.54 \\ \cline{2-8}
 & Best pair & 80.76,88.31) & (77.07,88.40) & (75.33,89.02) & (74.51,91.64) & (75.23,89.96) & (79.02,87.36) \\
 & GM & 84.45 & 82.54 & 81.89 & 82.63 & 82.27 & 83.09 \\ \hline
\multirow{4}{*}{SAM Output} & Sensitivity & 73.61$\pm$4.30 & 74.12$\pm$2.82 & 67.71$\pm$2.34 & 76.15$\pm$1.80 & 74.90$\pm$2.67 & 73.30$\pm$2.91 \\
 & Specificity & 92.68$\pm$1.85 & 92.60$\pm$1.31 & 92.03$\pm$1.95 & 90.44$\pm$1.46 & 90.64$\pm$2.48 & 91.55$\pm$1.75 \\ \cline{2-8}
 & Best pair & \textbf{(81.88,89.13)} & (80.35,90.46) & (70.11,91.70) & (75.33,91.66) & (75.58,91.31) & (76.04,91.33) \\
 & GM & \textbf{85.43} & 85.26 & 80.18 & 83.09 & 83.07 & 83.34 \\ \hline
\multirow{4}{*}{Zoom-in} & Sensitivity & 74.49$\pm$4.14 & 69.82$\pm$4.88 & 71.89$\pm$3.66 & 60.72$\pm$9.70 & 57.50$\pm$9.90 & 70.54$\pm$5.63 \\
 & Specificity & 92.48$\pm$2.62 & 92.32$\pm$2.47 & 91.53$\pm$1.77 & 93.32$\pm$3.21 & 94.39$\pm$2.31 & 89.75$\pm$3.11 \\ \cline{2-8}
 & Best pair & (84.03,86.05) & (74.71,90.78) & (77.17,89.59) & (72.36,90.16) & (70.42,90.62) & (83.11,82.54) \\
 & GM & 85.03 & 82.35 & 83.15 & 80.77 & 79.88 & 82.82 \\ \hline
\multirow{4}{*}{\begin{tabular}[c]{@{}l@{}}Geometric\\ (flip + rotation)\end{tabular}} & Sensitivity & 75.74$\pm$3.42 & 70.88$\pm$4.19 & 72.37$\pm$3.68 & 82.29$\pm$30.07 & 70.67$\pm$2.30 & 74.44$\pm$4.41 \\
 & Specificity & 92.26$\pm$1.92 & 92.43$\pm$1.83 & 91.88$\pm$1.78 & 18.72$\pm$30.46 & 94.72$\pm$0.69 & 93.05$\pm$1.94 \\ \cline{2-8}
 & Best pair & \textbf{(82.39,88.09)} & (79.63,89.43) & (80.45,87.81) & (60.29,49.82) & (73.29,93.97) & (81.37,88.58) \\
 & GM & \textbf{85.19} & 84.39 & 84.05 & 54.81 & 82.99 & 84.90 \\ \hline
\multirow{4}{*}{Colour Jitter} & Sensitivity & 74.77$\pm$2.84 & 65.42$\pm$6.96 & 67.38$\pm$4.86 & 58.52$\pm$47.84 & 67.87$\pm$40.64 & 60.16$\pm$11.89 \\
 & Specificity & 91.57$\pm$1.51 & 91.87$\pm$1.38 & 92.59$\pm$1.86 & 41.36$\pm$47.96 & 34.58$\pm$40.05 & 89.72$\pm$8.97 \\ \cline{2-8}
 & Best pair & (77.99,91.31) & (72.36,89.75) & (75.23,90.06) & (92.22,9.48) & (80.25,43.48) & (76.66,84.67) \\
 & GM & 84.39 & 80.59 & 82.31 & 29.57 & 59.07 & 80.57 \\ \hline
\end{tabular}
\end{table*}

We systematically evaluated the influence of four data augmentation strategies(segmentation-based, zoom-in, geometric transformations, and colour jitter) on model performance (Table \ref{augmentation_performance_all_models}). Geometric transformations, specifically horizontal-vertical flips combined with random rotations, emerged as the most effective strategy, yielding consistent improvements in both sensitivity and specificity across all tested lightweight architectures. 

In contrast, photometric and segmentation-based augmentations produced variable or detrimental results. Colour jitter actively degraded metrics for NASNet-Mobile, EfficientNetV2-B0, and MobileNetV3-Large. This suggests that aggressive chromatic modification obscures critical diagnostic features (e.g., red vs. white lesions). Similarly, zoom-in augmentations showed inconsistent impacts, indicating that loss of contextual texture hampers the receptive fields of certain architectures.

Finally, while segmentation-guided augmentation (using SAM) theoretically improves focus, a qualitative review revealed that it frequently produced erroneous masks due to the lack of precise bounding boxes. Also, from a systems-engineering perspective, excluding SAM is advantageous: it removes a computationally intensive, high-memory dependency from the data curation pipeline, streamlining the training process without compromising diagnostic accuracy.

\subsection{Reweighting based on Image Quality}
\label{reweighting}

\begin{table*}[htbp]
\caption{Performance of models with image-quality based reweighting of loss functions}
\label{reweighting_results}
\centering
\renewcommand{\arraystretch}{1.25}
\begin{tabular}{|l|l|c|c|c|c|c|c|}
\hline
\textbf{Method} & \textbf{Metric} & \textbf{MobileViTv2} & \textbf{EdgeNeXt-S} & \textbf{DeiT-Ti} & \textbf{MobileNetV3-L} & \textbf{EffNetV2-B0} & \textbf{NASNet-Mobile} \\ \hline
\multirow{4}{*}{Baseline} & Sensitivity & 75.48$\pm$3.69 & 72.01$\pm$3.48 & 69.39$\pm$4.25 & 59.48$\pm$12.46 & 61.03$\pm$8.50 & 72.11$\pm$3.85 \\
 & Specificity & 90.45$\pm$3.03 & 90.42$\pm$2.61 & 91.52$\pm$1.68 & 93.99$\pm$2.61 & 94.91$\pm$2.50 & 90.95$\pm$2.54 \\ \cline{2-8}
 & Best pair & \textbf{(80.76,88.31)} & (77.07,88.40) & (75.33,89.02) & (74.51,91.64) & (75.23,89.96) & (79.02,87.36) \\
 & GM & \textbf{84.45} & 82.54 & 81.89 & 82.63 & 82.27 & 83.09 \\ \hline
\multirow{4}{*}{\begin{tabular}[c]{@{}l@{}}NIQE\\ Reweighting\end{tabular}} & Sensitivity & 69.90$\pm$3.96 & 63.21$\pm$8.14 & 65.14$\pm$5.15 & 65.12$\pm$5.60 & 62.28$\pm$3.28 & 64.38$\pm$4.13 \\
 & Specificity & 92.15$\pm$2.75 & 90.60$\pm$3.05 & 91.68$\pm$4.48 & 91.39$\pm$2.60 & 92.28$\pm$4.50 & 91.37$\pm$1.40 \\ \cline{2-8}
 & Best pair & (72.13,91.50) & (66.76,90.18) & (65.25,91.28) & (66.28,90.12) & (63.15,91.28) & (65.19,90.82) \\
 & GM & 81.24 & 77.59 & 77.18 & 77.29 & 75.92 & 76.95 \\ \hline
\multirow{4}{*}{\begin{tabular}[c]{@{}l@{}}CLIP-IQA\\ Reweighting\end{tabular}} & Sensitivity & 69.65$\pm$3.83 & 63.41$\pm$6.54 & 65.04$\pm$4.68 & 65.14$\pm$5.15 & 62.32$\pm$3.98 & 63.68$\pm$4.33 \\
 & Specificity & 93.25$\pm$2.86 & 90.60$\pm$3.86 & 91.74$\pm$4.21 & 91.68$\pm$2.46 & 92.37$\pm$4.56 & 90.73$\pm$1.43 \\ \cline{2-8}
 & Best pair & (71.68,91.80) & (66.46,90.82) & (67.17,89.40) & (65.17,90.47) & (64.45,91.12) & (64.19,90.42) \\
 & GM & \ 81.12 & 77.69 & 77.49 & 76.78 & 76.63 & 76.18 \\ \hline
\end{tabular}
\end{table*}

Reweighting loss functions based on no-reference IQA scores (NIQE and CLIP-IQA) yielded performance largely comparable to the baseline, offering no significant overall advantage (Table \ref{reweighting_results}). The trade-offs varied by architecture, with EdgeNeXt-S, DeiT-Ti, and NasNet-Mobile experiencing a reduction in average sensitivity with no compensatory gain in specificity. 

These findings suggest a fundamental misalignment between generic image quality metrics and diagnostic utility: emphasising images with high perceptual aesthetic scores may inadvertently down-weight noisy samples that may yet contain good discriminative clinical features. Given that the training data was collected in the field, images are expected to contain some noise, potentially leading to downweighting a large proportion of the good training examples. Further, integrating complex perceptual IQA models adds unnecessary computational overhead to the training pipeline without yielding significant downstream performance gains. 

\subsection{Strategies for Handling Class Imbalance}
\label{class_imbalance}

\begin{table*}[htbp]
\centering
\caption{Performance of models with various techniques to handle class imbalance}
\label{tab: class_imbalance_results}
\renewcommand{\arraystretch}{1.2}
\footnotesize
\begin{tabular}{|l|l|c|c|c|c|c|c|}
\hline
\textbf{Method} & \textbf{Metric} & \textbf{MobileViTv2} & \textbf{EdgeNeXt-S} & \textbf{DeiT-Ti} & \textbf{MobileNetV3-L} & \textbf{EffNetV2-B0} & \textbf{NASNet-Mobile} \\ \hline
\multirow{4}{*}{Baseline} & Sensitivity & 75.48$\pm$3.69 & 72.01$\pm$3.48 & 69.39$\pm$4.25 & 59.48$\pm$12.46 & 61.03$\pm$8.50 & 72.11$\pm$3.85 \\
 & Specificity & 90.45$\pm$3.03 & 90.42$\pm$2.61 & 91.52$\pm$1.68 & 93.99$\pm$2.61 & 94.91$\pm$2.50 & 90.95$\pm$2.54 \\ \cline{2-8}
 & Best pair & (80.76,88.31) & (77.07,88.40) & (75.33,89.02) & (74.51,91.64) & (75.23,89.96) & (79.02,87.36) \\
 & GM & 84.45 & 82.54 & 81.89 & 82.63 & 82.27 & 83.09 \\ \hline
\multirow{4}{*}{\begin{tabular}[c]{@{}l@{}}Random\\ Undersampling\end{tabular}} & Sensitivity & 83.63$\pm$3.60 & 82.03$\pm$2.54 & 83.13$\pm$2.87 & 82.57$\pm$3.29 & 82.82$\pm$11.71 & 81.42$\pm$3.09 \\
 & Specificity & 85.17$\pm$3.66 & 85.54$\pm$2.23 & 82.88$\pm$2.46 & 81.32$\pm$2.42 & 77.86$\pm$7.29 & 81.71$\pm$4.73 \\ \cline{2-8}
 & Best pair & (87.82,82.66) & (82.09,87.65) & (83.11,85.32) & (84.65,81.75) & (83.73,81.71) & (83.93,81.71) \\
 & GM & 85.20 & 84.82 & 84.21 & 83.19 & 82.71 & 82.81 \\ \hline
\multirow{4}{*}{\begin{tabular}[c]{@{}l@{}}Random\\ Oversampling\end{tabular}} & Sensitivity & 76.21$\pm$3.72 & 84.34$\pm$5.07 & 74.49$\pm$4.81 & 69.23$\pm$9.18 & 77.41$\pm$4.88 & 69.61$\pm$3.69 \\
 & Specificity & 90.10$\pm$3.06 & 81.52$\pm$5.11 & 88.55$\pm$3.45 & 88.76$\pm$7.23 & 88.70$\pm$3.41 & 91.33$\pm$3.67 \\ \cline{2-8}
 & Best pair & (77.69,90.76) & (87.51,81.02) & (82.29,83.05) & (81.47,82.58) & (81.27,88.25) & (71.24,91.54) \\
 & GM & 83.97 & 84.20 & 82.67 & 82.02 & 84.69 & 80.75 \\ \hline
\multirow{4}{*}{\begin{tabular}[c]{@{}l@{}}Weighted Cross\\ Entropy\end{tabular}} & Sensitivity & 83.79$\pm$2.85 & 82.98$\pm$1.71 & 84.23$\pm$2.83 & 77.35$\pm$6.44 & 82.91$\pm$3.82 & 73.93$\pm$7.19 \\
 & Specificity & 85.71$\pm$3.10 & 84.40$\pm$2.16 & 82.59$\pm$3.30 & 88.02$\pm$6.42 & 82.69$\pm$4.59 & 86.98$\pm$7.41 \\ \cline{2-8}
 & Best pair & \textbf{(88.54,84.26)} & (83.42,85.95) & (85.57,82.91) & (74.51,91.64) & (79.32,87.55) & (79.84,85.40) \\
 & GM & \textbf{86.37} & 84.68 & 84.23 & 82.63 & 83.33 & 82.57 \\ \hline
\multirow{4}{*}{Focal Loss} & Sensitivity & 82.19$\pm$2.92 & 83.37$\pm$3.41 & 80.85$\pm$2.65 & 64.85$\pm$6.67 & 62.58$\pm$9.36 & 64.85$\pm$6.67 \\
 & Specificity & 87.92$\pm$2.68 & 87.09$\pm$2.50 & 84.22$\pm$2.19 & 94.75$\pm$1.49 & 94.45$\pm$2.86 & 94.75$\pm$1.49 \\ \cline{2-8}
 & Best pair & (81.17,90.99) & (88.02,84.59) & (83.73,83.21) & (74.31,92.47) & (79.02,88.09) & (74.31,92.47) \\
 & GM & 85.94 & 86.29 & 83.47 & 82.89 & 83.43 & 82.89 \\ \hline
\multirow{4}{*}{\begin{tabular}[c]{@{}l@{}}Balanced Mix-up\\ ($\alpha=0.1$)\end{tabular}} & Sensitivity & 87.63$\pm$4.48 & 92.07$\pm$4.42 & 78.00$\pm$7.61 & 80.24$\pm$7.55 & 88.50$\pm$3.39 & 80.42$\pm$7.74 \\
 & Specificity & 63.19$\pm$2.66 & 59.43$\pm$9.48 & 71.59$\pm$9.17 & 58.40$\pm$8.87 & 53.00$\pm$5.83 & 60.87$\pm$6.90 \\ \cline{2-8}
 & Best pair & (84.81,73.49) & (86.39,73.39) & (78.33,77.79) & (75.17,66.60) & (86.72,63.15) & (74.89,71.80) \\
 & GM & 78.95 & 79.63 & 78.06 & 70.76 & 74.00 & 73.33 \\ \hline
\multirow{4}{*}{\begin{tabular}[c]{@{}l@{}}Balanced Mix-up\\ ($\alpha=0.2$)\end{tabular}} & Sensitivity & 83.94$\pm$8.55 & 92.08$\pm$2.85 & 85.01$\pm$6.67 & 79.36$\pm$5.55 & 88.87$\pm$3.80 & 79.25$\pm$7.84 \\
 & Specificity & 67.24$\pm$9.57 & 59.27$\pm$6.52 & 63.55$\pm$11.41 & 58.94$\pm$8.74 & 53.37$\pm$5.15 & 60.51$\pm$6.88 \\ \cline{2-8}
 & Best pair & (83.94,74.82) & (86.35,72.67) & (78.64,75.44) & (78.41,67.69) & (81.11,62.54) & (75.15,71.91) \\
 & GM & 79.25 & 79.22 & 77.02 & 72.85 & 71.22 & 73.51 \\ \hline
\multirow{4}{*}{\begin{tabular}[c]{@{}l@{}}Balanced Mix-up\\ ($\alpha=0.3$)\end{tabular}} & Sensitivity & 81.98$\pm$8.54 & 92.23$\pm$3.04 & 81.83$\pm$4.10 & 75.72$\pm$7.36 & 88.75$\pm$4.95 & 79.15$\pm$7.74 \\
 & Specificity & 68.00$\pm$10.50 & 59.73$\pm$5.90 & 67.69$\pm$6.72 & 60.88$\pm$7.94 & 53.25$\pm$7.26 & 60.71$\pm$6.89 \\ \cline{2-8}
 & Best pair & (74.09,79.32) & (84.25,60.35) & (75.11,77.89) & (79.45,66.83) & (88.13,60.29) & (75.01,72.04) \\
 & GM & 76.66 & 71.31 & 76.49 & 72.87 & 72.89 & 73.51 \\ \hline
\end{tabular}
\end{table*}

As detailed in Table \ref{tab: class_imbalance_results}, among the sampling-based strategies, random undersampling yielded the most significant increase in sensitivity, effectively targeting the minority class, but at the expense of a slight reduction in specificity. Conversely, random oversampling offered a more conservative improvement but increased the overall training burden by artificially inflating the dataset size.

Among the loss-based strategies, Weighted Cross-Entropy (WCE) emerged as the most robust and efficient approach. By weighting the loss function based on the inverse of class frequency, WCE consistently improved sensitivity across all models. From an engineering standpoint, WCE introduces no additional computational overhead during training (unlike oversampling) and maintains the integrity of the original data distribution (unlike undersampling or Mix-up). 

Focal Loss exhibited architectural instability, causing performance inversions in MobileNetV3-L, EfficientNetV2-B0, and NASNet-Mobile. Balanced Mix-up boosted sensitivity but caused a precipitous drop in specificity, resulting in a significant increase in false positives (undesirable for a triaging system). Thus, WCE provided a superior equilibrium, offering substantial gains in minority class detection while maintaining overall system reliability and computational efficiency.

\subsection{Final Model Architecture Selection and Five-fold Cross Validation}
Synthesising the ablation findings, the combination of geometric augmentations and Weighted Cross-Entropy loss was identified as the most efficient training strategy. Architecturally, MobileViTv2 emerged as the most robust candidate with consistent performance across experimental setups. Pure CNNs like EfficientNetV2-B0 and MobileNetV3-Large proved highly sensitive to hyperparameter variations, and NASNet-Mobile exhibited persistent underperformance across various experiments. The transformer-based and hybrid models, DeiT-Ti and EdgeNeXt-S, exhibited slightly lower performance than MobileViT-V2, suggesting that MobileViT-V2, combined with geometric augmentations and Weighted Cross-Entropy loss, is the optimal training paradigm.

MobileViTv2's success highlights the advantage of hybrid architectures for edge deployment: it leverages the parameter efficiency and local feature extraction of convolutions alongside the global context modelling of transformers, all within a constrained $\sim$4.9M parameter footprint. 

\begin{table}[H] 
\caption{Five-fold cross-validation of MobileViTv2 architecture trained with the optimal strategy}
\label{crossval}
\resizebox{\linewidth}{!}{%
\begin{tabular}{|c|c|c|c|c|}
\hline
\textbf{Fold} & \textbf{Sensitivity (\%)} & \textbf{Specificity (\%)} & \textbf{Best Pair (\%)} \\
\hline
\textbf{1}  & 84.37 ± 2.65 & 85.81 ± 2.41 & 87.41, 86.49 \\
\textbf{2} & 82.99 ± 3.51 & 86.20 ± 3.47 & 83.21, 86.98 \\
\textbf{3} & 84.08 ± 3.55 & 85.02 ± 4.12 & 83.44, 86.81 \\
\textbf{4} & 80.46 ± 2.94 & 86.99 ± 3.16 & 83.95, 85.62 \\
\textbf{5} & 84.23 ± 1.82 & 86.22 ± 2.16 & 85.16, 86.17 \\
\hline
\end{tabular}%
}
\end{table}

Further, a five-fold cross-validation of the MobileViTv2 architecture trained with the optimal strategy of geometric augmentations and Weighted Cross-Entropy loss confirmed systemic stability as shown in Table \ref{crossval}. The models exhibited a mean sensitivity of 83.2 $\pm$ 1.5\% and a mean specificity of 86.0 $\pm$ 0.8\%. The low standard deviation indicates robust generalisation, reducing concerns of overfitting to a particular fold. The best metrics were achieved by the best model for fold-1, with a sensitivity of 87.41\% and a specificity of 86.49\%. Additionally, this model had a high negative predictive value of 97.2\%, making it suitable for screening and establishing the viability of the proposed lightweight architecture for scalable edge deployments.


\subsection{Knowledge Distillation Ablation}
\label{distillation}

To determine if the performance of the lightweight MobileViTv2 model could be enhanced without increasing its inference latency, we evaluated Knowledge Distillation (KD) \cite{hinton2015distilling}. This technique attempts to transfer complex feature representations from high-capacity, compute-intensive ``teacher'' models to a smaller ``student'' model. 

We initially trained three resource-intensive architectures, Vision Transformer, ViT-B/16 with approximately $86$M parameters \cite{dosovitskiy2021an}, EfficientNetV2-L with approximately $ 118$M parameters \cite{tan2021efficientnetv2}, and Swin Transformer, Swin-B with approximately $ 88$M parameters \cite{liu2021swin}, as teacher networks using weighted cross-entropy loss and geometric augmentations. While these large models achieved baseline sensitivities between 79.9\% and 84.3\% and specificities near 89\%, their massive parameter counts and FLOP requirements render them unsuitable for edge deployment in resource-constrained environments. 

We subsequently performed distillation onto the MobileViTv2 student model, modulating the KD loss contribution relative to the ground-truth loss. As detailed in Table \ref{tab:combined_kd}, the distillation ablation revealed several critical pipeline inefficiencies that ultimately disqualified KD from our proposed system:

\begin{itemize}
    \item \textbf{Training Instability and Model Collapse:} Distillation proved highly unstable under specific configurations. Notably, relying entirely on the EfficientNetV2-L teacher (100\% KD contribution) induced catastrophic model collapse, dropping average sensitivity to 38.16 $\pm$ 32.92\%. We attribute this mainly to an architectural inductive bias mismatch and partly to a capacity mismatch. EfficientNetV2-L has $\sim 30$M additional parameters than Swin-B and ViT-B/16. Moreover, EfficientNetV2-L is a pure CNN-based model, while MobileViTv2 is a hybrid model and relies heavily on transformer blocks in addition to CNN layers. Forcing the student to mimic pure CNN-like soft-label distributions derived from local feature representations, without ground-truth supervision, renders the hybrid representation of MobileViTv2 incompatible with its teacher's inductive biases. In contrast, distillation from the transformer-based ViT-B/16 and Swin-B teachers remained relatively stable at 100\% distillation, with distillation from ViT-B/16 producing the best results, highlighting that distillation from a similar-architecture model is the optimal strategy.
    
    \item \textbf{Decision Boundary Bias:} Using the Swin-B at 100\% distillation improved sensitivity to 86.43 $\pm$ 1.85\% but collapsed specificity to 62.65 $\pm$ 2.63\%. Without ground-truth supervision, the student over-commits to the teacher's soft-label distribution and shifts its decision boundary toward the suspicious class, sacrificing specificity. The 50\% configuration restores this sensitivity/specificity balance (83.50\% / 87.37\%), confirming that blended hard- and soft-label supervision is necessary to anchor the student's calibration.

    \item \textbf{Marginal Gains vs. Compute Overhead:} Although distillation from ViT-B/16 (50\% KD loss contribution) yielded competitive metrics (84.22 $\pm$ 3.56\% sensitivity, 87.37 $\pm$ 3.53\% specificity), it failed to strictly outperform the baseline MobileViTv2 trained directly on ground-truth labels. This marginal improvement and variance around the estimate does not justify the massive computational and memory overhead required to train, store, and infer from a teacher model during the training pipeline.
\end{itemize}

Given that KD introduced severe pipeline complexity, convergence instability, and capacity bottlenecks without yielding robust performance improvements, we excluded it from the final deployment framework in favour of direct training of lightweight models on ground truth labels.

\begin{table}[htbp]
\caption{Distillation Ablation: Performance of MobileViTv2 (Student) distilled from high-capacity teachers. For reference, the non-distilled MobileViTv2 baseline achieves 75.48\% sensitivity and 90.45\% specificity (best pair: 80.76\%, 88.31\%).}
\label{tab:combined_kd}
\centering
\renewcommand{\arraystretch}{1.2}
\resizebox{\linewidth}{!}{%
\begin{tabular}{|>{\centering\arraybackslash}m{2.1cm}|>{\centering\arraybackslash}m{1.5cm}|>{\centering\arraybackslash}m{2cm}|>{\centering\arraybackslash}m{2cm}|>{\centering\arraybackslash}m{2.5cm}|}
\hline
\textbf{Teacher Model (Baseline Sens, Spec)} & \textbf{KD Loss (\%)} & \textbf{Student Sensitivity (\%)} & \textbf{Student Specificity (\%)} & \textbf{Student Best Pair (Sens, Spec) (\%)} \\ \hline
\multicolumn{2}{|c|}{\textbf{MobileViTv2 baseline (no KD)}} & 75.48 $\pm$ 3.69 & 90.45 $\pm$ 3.03 & 80.76, 88.31 \\ \hline
\multirow{2}{*}{\begin{tabular}[c]{@{}l@{}}ViT-B/16\\ (79.94\%, 89.51\%)\end{tabular}} & 50 & 84.22 $\pm$ 3.56 & 87.37 $\pm$ 3.53 & 85.47, 88.64 \\
 & 100 & 79.55 $\pm$ 3.37 & 89.71 $\pm$ 2.65 & 85.47, 85.18 \\ \hline
\multirow{2}{*}{\begin{tabular}[c]{@{}l@{}}EfficientNetV2-L\\ (84.34\%, 88.92\%)\end{tabular}} & 50 & 82.44 $\pm$ 2.64 & 86.33 $\pm$ 2.83 & 87.72, 82.97 \\
 & 100 & 38.16 $\pm$ 32.92 & 62.83 $\pm$ 31.86 & 55.89, 56.38 \\ \hline
\multirow{2}{*}{\begin{tabular}[c]{@{}l@{}}Swin-B\\ (80.24\%, 88.88\%)\end{tabular}} & 50 & 83.50 $\pm$ 3.05 & 87.37 $\pm$ 3.12 & 82.59, 89.48 \\
 & 100 & 86.43 $\pm$ 1.85 & 62.65 $\pm$ 2.63 & 83.52, 66.92 \\ \hline
\end{tabular}%
}
\end{table}

\section{Interpretability Validation}
\label{interpretability}

In unconstrained edge deployments where images are captured using heterogeneous smartphone hardware, deep learning models are highly susceptible to learning spurious correlations, such as sensor noise, flash reflections, or background artefacts, rather than true pathological features. To validate that our optimised lightweight system derives its predictions from robust, clinically relevant signals rather than acquisition artifacts, we conducted an interpretability analysis on the selected MobileViTv2 model. 

Because MobileViTv2 is a hybrid architecture, we employed a dual-technique approach to visualise its interpretability: Gradient-weighted Class Activation Mapping (GradCAM++) \cite{gradcampp} to evaluate the local feature extraction of the convolutional layers, and Attention Rollout \cite{abnar2020quantifying} to map the global context captured by the Transformer blocks at various spatial scales. To align the interpretability analysis directly with the network's intrinsic hierarchical design, we extracted attention maps at the specific patch scales utilised by MobileViTv2 (16, 64, and 256). These scales correspond to the allocation of 3, 4, and 2 self-attention blocks at each respective stage of feature extraction, allowing us to track how the system aggregates information from granular local patches up to global structural contexts.

\begin{figure}[htbp]
    \centering
    \begin{subfigure}[t]{0.32\linewidth}
        \centering
        \includegraphics[width=\linewidth]{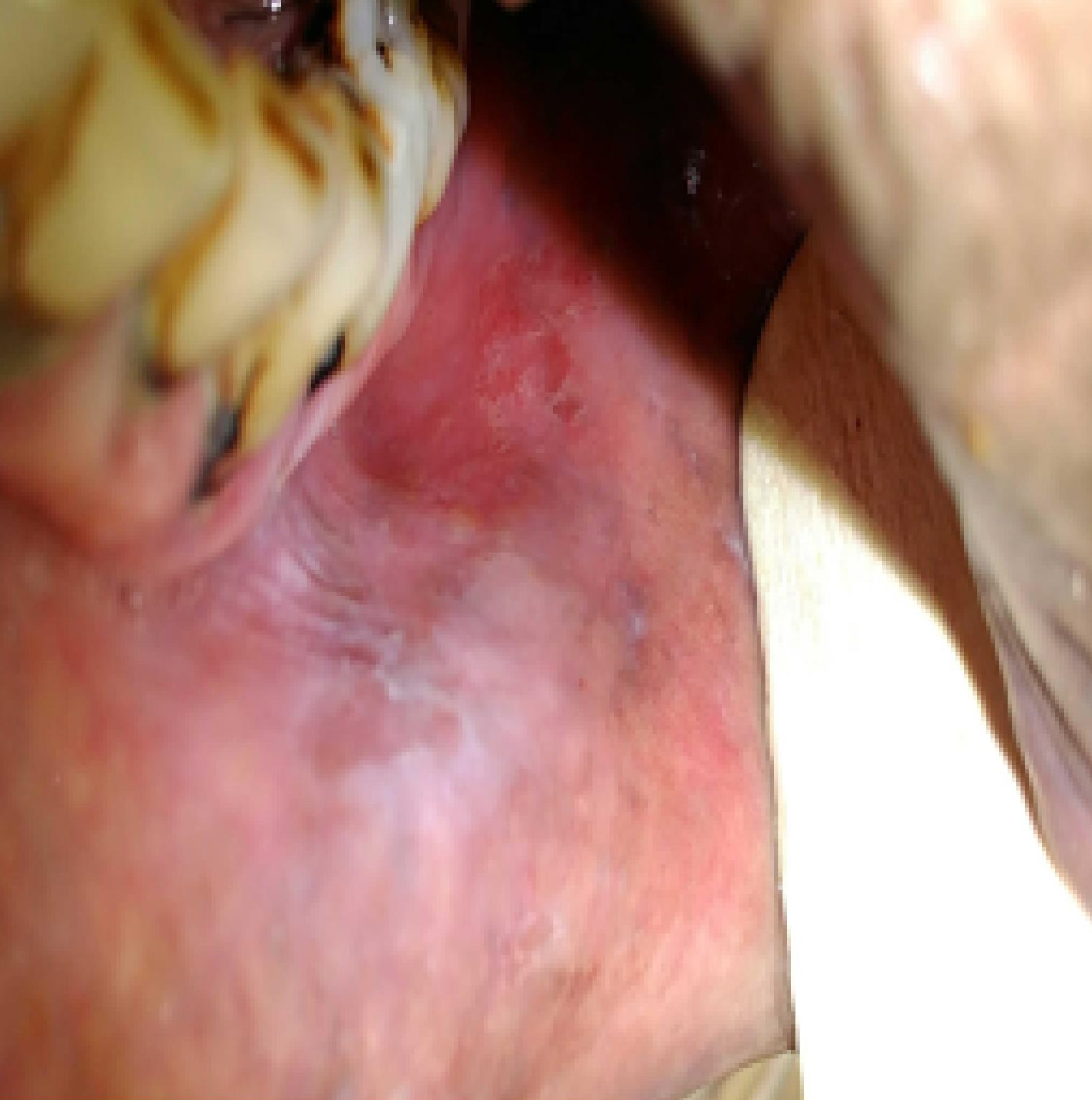}
        \caption{Original}
        \label{gradcam_org}
    \end{subfigure}
    \hfill
    \begin{subfigure}[t]{0.32\linewidth}
        \centering
        \includegraphics[width=\linewidth]{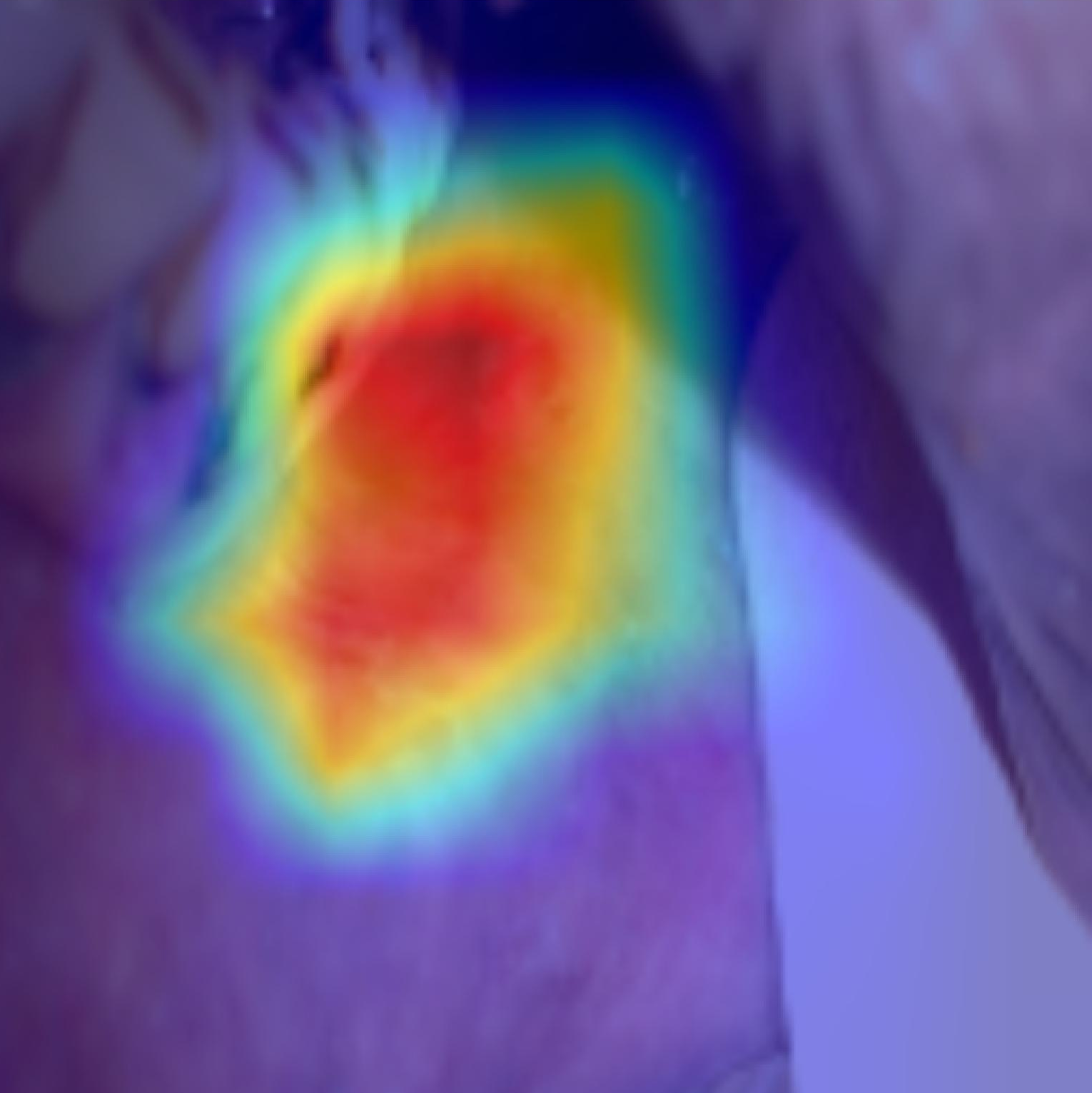}
        \caption{GradCAM++ o/p}
        \label{gradcam_heatmap}
    \end{subfigure}
    \hfill
    \begin{subfigure}[t]{0.32\linewidth}
        \centering
        \includegraphics[width=\linewidth]{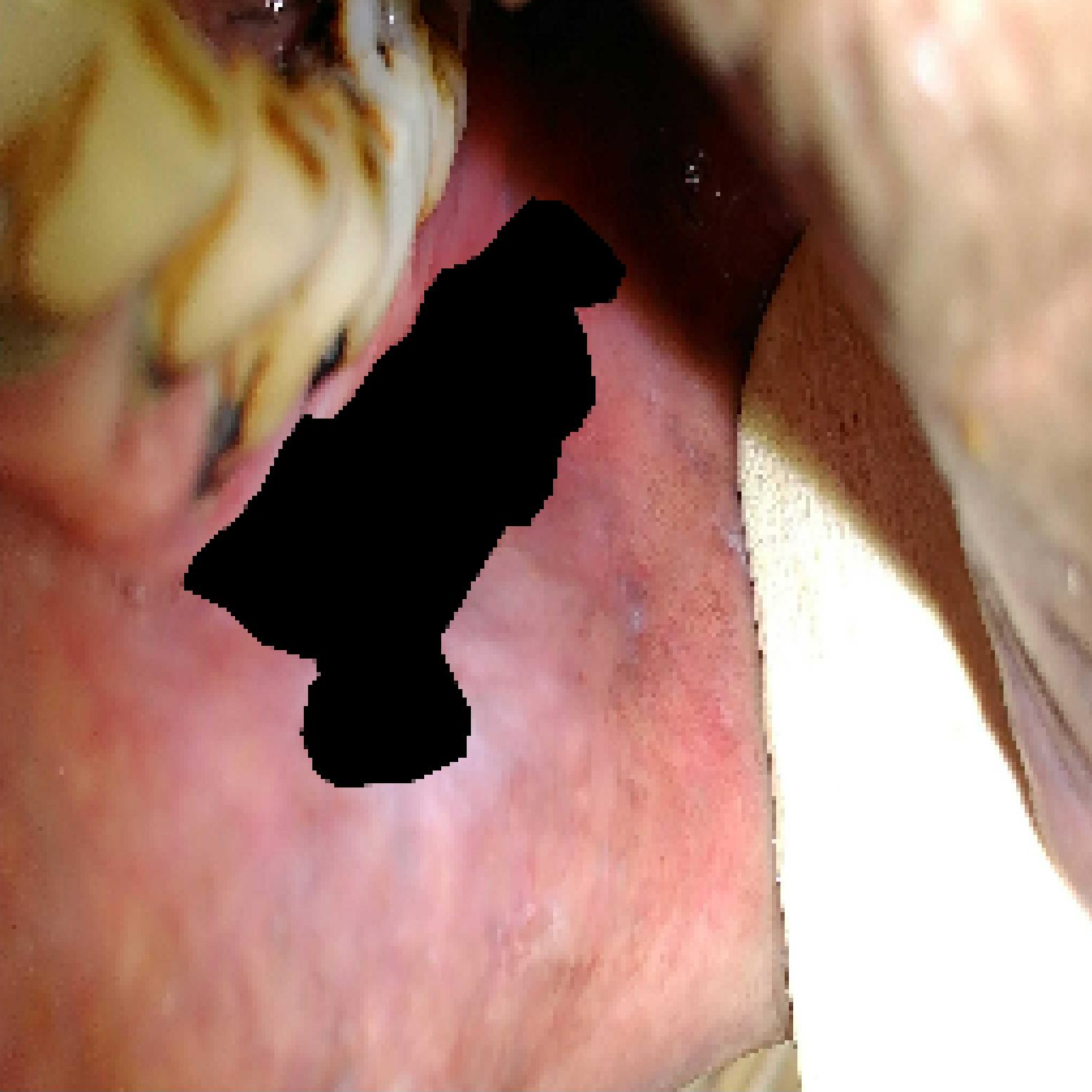}
        \caption{Expert annotation}
        \label{gradcam_annotation}
    \end{subfigure}
    \caption{Interpretability via Grad-CAM++. The highest model activations in (b) closely align with the specialist's or expert's lesion annotation in (c), confirming that the model avoids spurious background correlations.}
    \label{TP_nhl}
\end{figure}

\begin{figure}[htbp]
    \centering
    \begin{subfigure}[t]{0.24\linewidth}
        \centering
        \includegraphics[width=\linewidth]{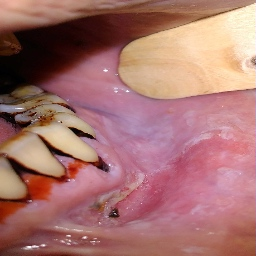}
        \caption{Original}
        \label{attn_org}
    \end{subfigure}
    \hfill
    \begin{subfigure}[t]{0.24\linewidth}
        \centering
        \includegraphics[width=\linewidth]{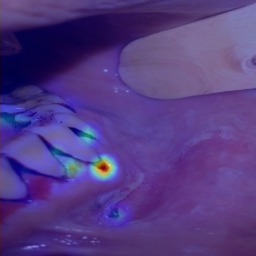}
        \caption{Scale: 256}
        \label{attn_256}
    \end{subfigure}
    \hfill
    \begin{subfigure}[t]{0.24\linewidth}
        \centering
        \includegraphics[width=\linewidth]{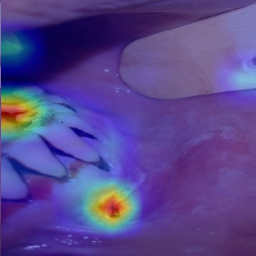}
        \caption{Scale: 64}
        \label{attn_64}
    \end{subfigure}
    \hfill
    \begin{subfigure}[t]{0.24\linewidth}
        \centering
        \includegraphics[width=\linewidth]{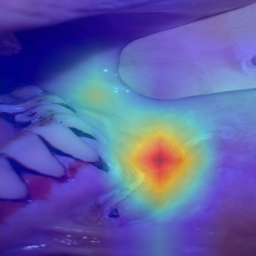}
        \caption{Scale: 16}
        \label{attn_16}
    \end{subfigure}
    \caption{Multi-scale Transformer Attention Rollout for a true-positive Suspicious classification, demonstrating the model's extraction of global context at feature scales 256, 64, and 16.}
    \label{fig:attention-sus}
\end{figure}

As shown in Figure \ref{TP_nhl}, the GradCAM++ heatmaps generated by our system exhibit tight spatial alignment with ground-truth segmentations provided by clinical specialists. Further, the Transformer attention maps (Figure \ref{fig:attention-sus}) confirm that the model successfully aggregates multi-scale contextual features from the oral cavity to inform its inference.

\section{Robustness and Stress Testing}
\label{robustness}

In resource-constrained edge environments, deep learning models must remain robust not only to anatomical variations but also to hardware-induced signal degradation. To stress-test the operational limits of the best-performing MobileViTv2 model, we evaluated its robustness against three common types of image noise, mapping them to real-world acquisition and hardware failures: unstructured Gaussian noise (simulating low-light sensor noise), structured stripe noise (simulating sensor read-out artefacts), and salt-and-pepper (impulse) noise (simulating stuck pixels or transmission bit-errors) \cite{ghafari2021robustness}. 

\begin{figure}[htbp]
    \centering
    \begin{subfigure}[t]{0.24\linewidth}
        \centering
        \includegraphics[width=\linewidth]{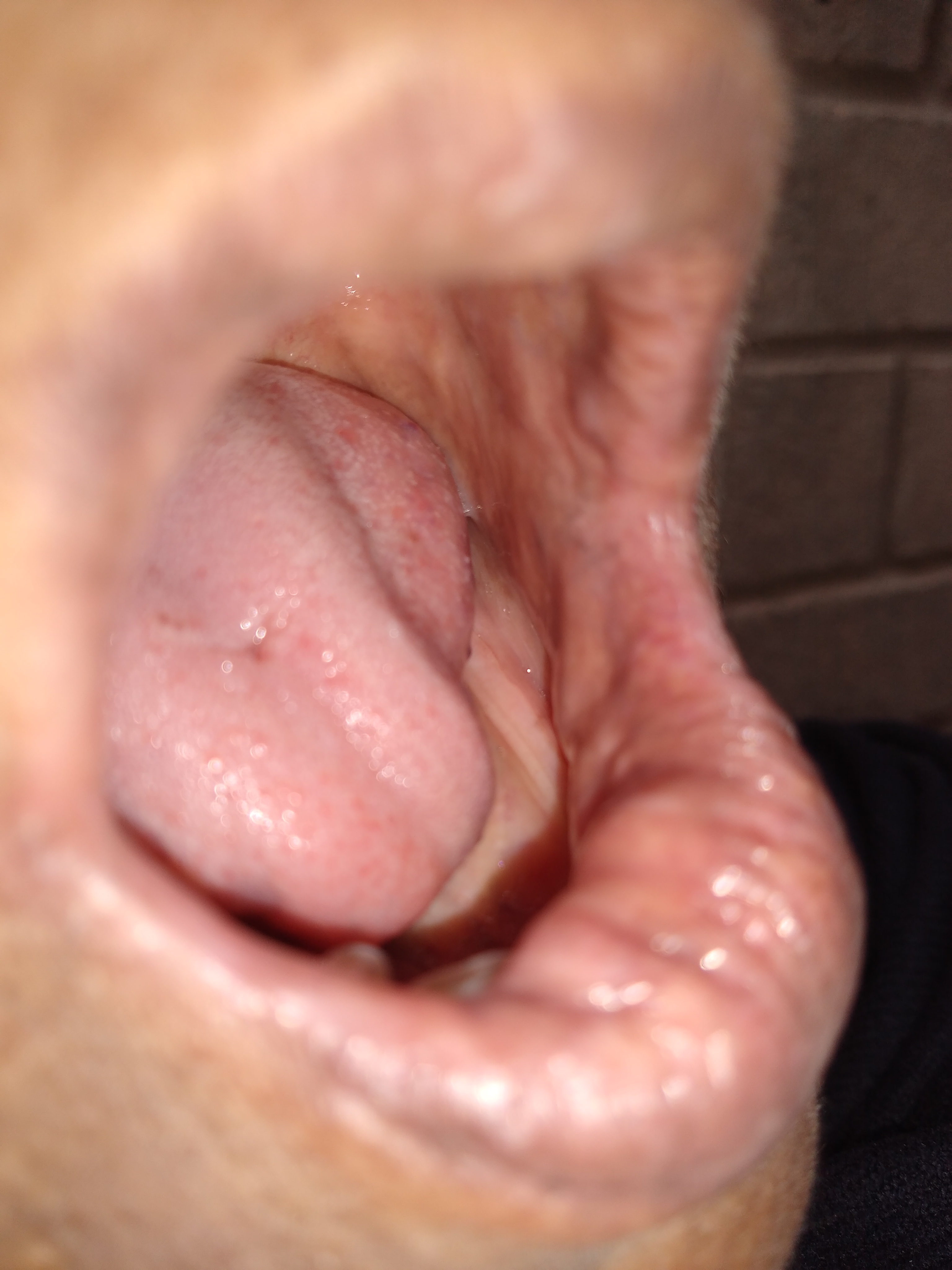}
        \caption{Original}
    \end{subfigure}
    \hfill
    \begin{subfigure}[t]{0.24\linewidth}
        \centering
        \includegraphics[width=\linewidth]{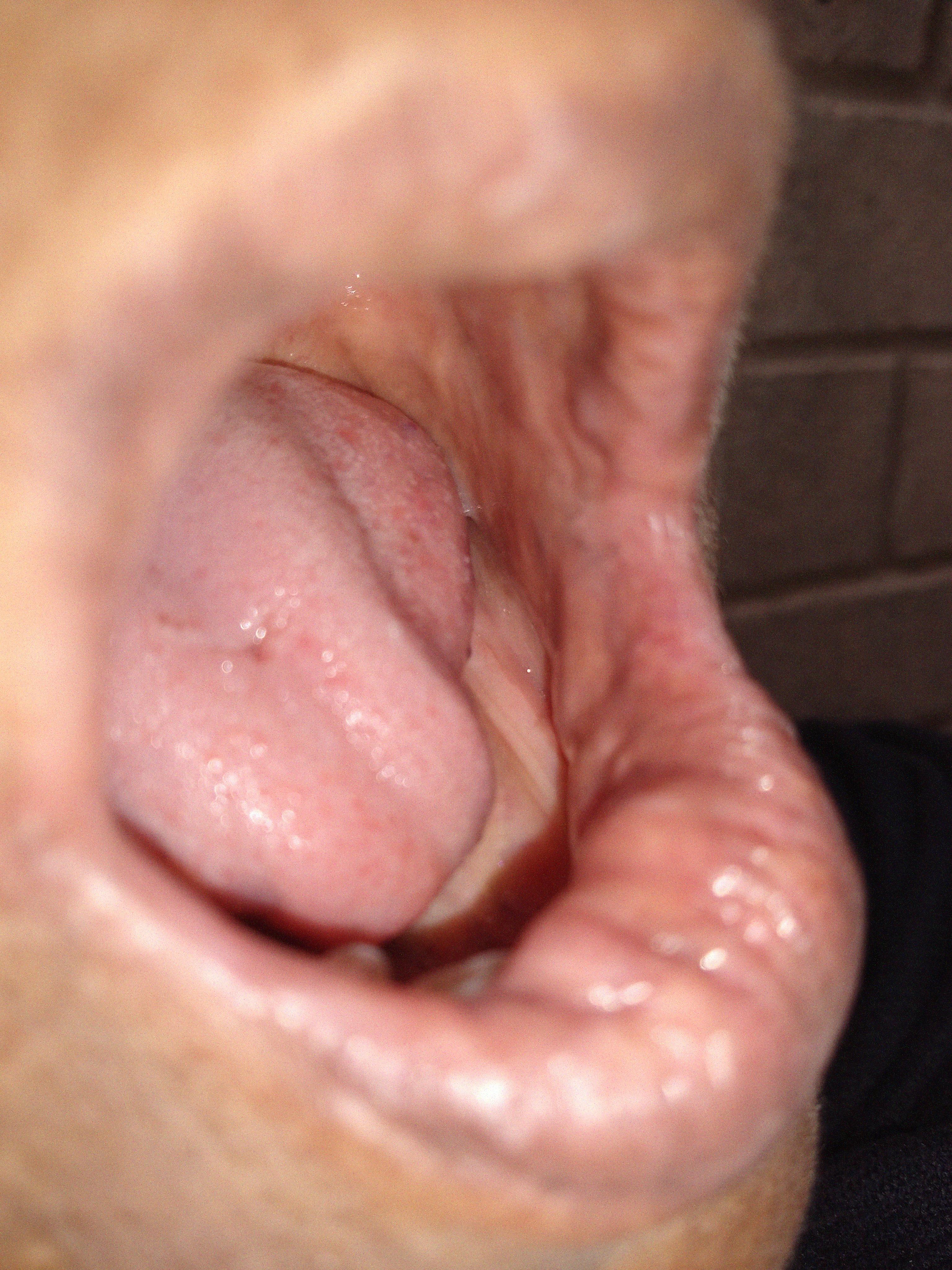}
        \caption{Gauss (20\%)}
        \label{gaussian}
    \end{subfigure}
    \hfill
    \begin{subfigure}[t]{0.24\linewidth}
        \centering
        \includegraphics[width=\linewidth]{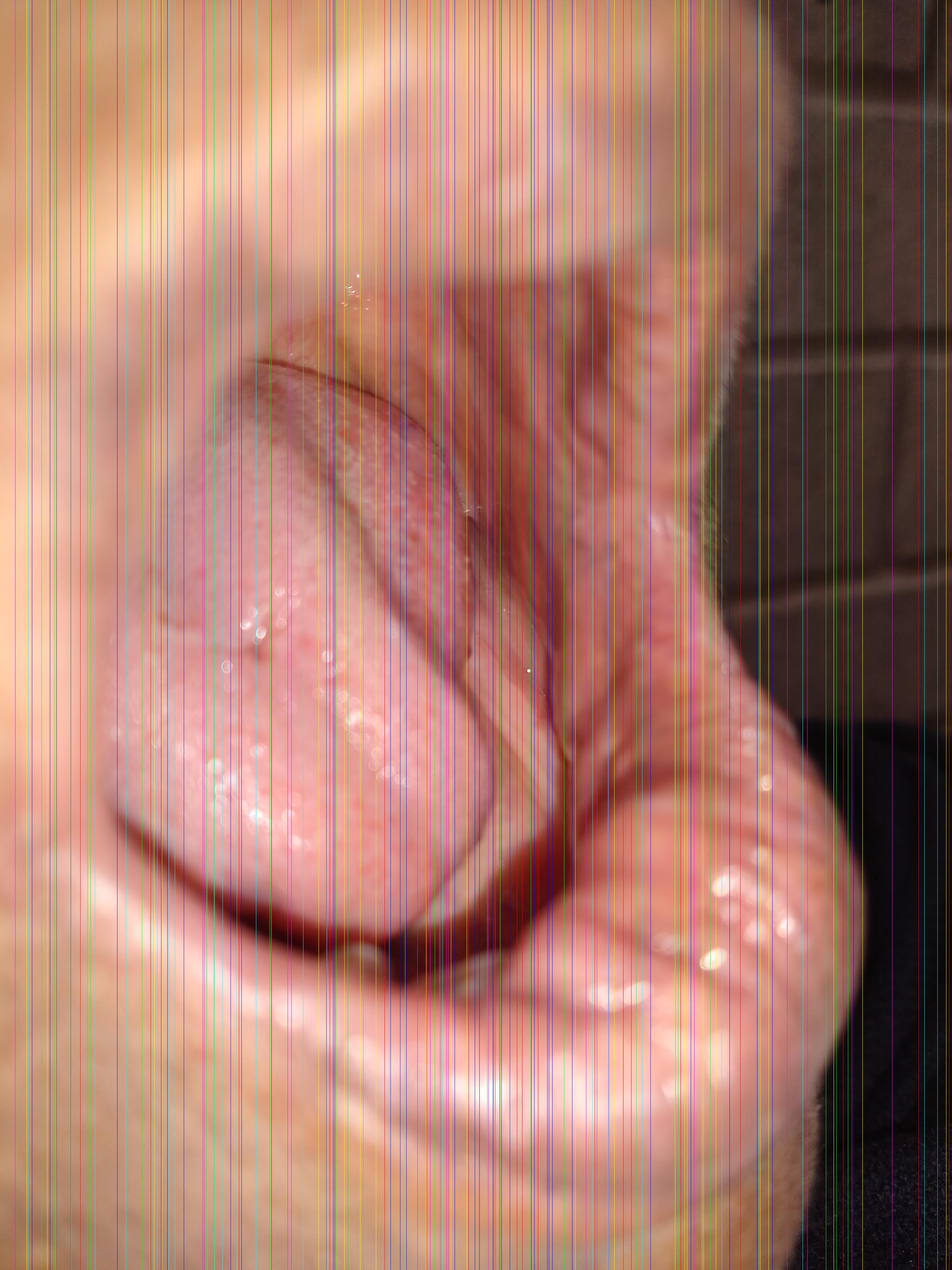}
        \caption{Stripe (5\%)}
        \label{stripe}
    \end{subfigure}
    \hfill
    \begin{subfigure}[t]{0.24\linewidth}
        \centering
        \includegraphics[width=\linewidth]{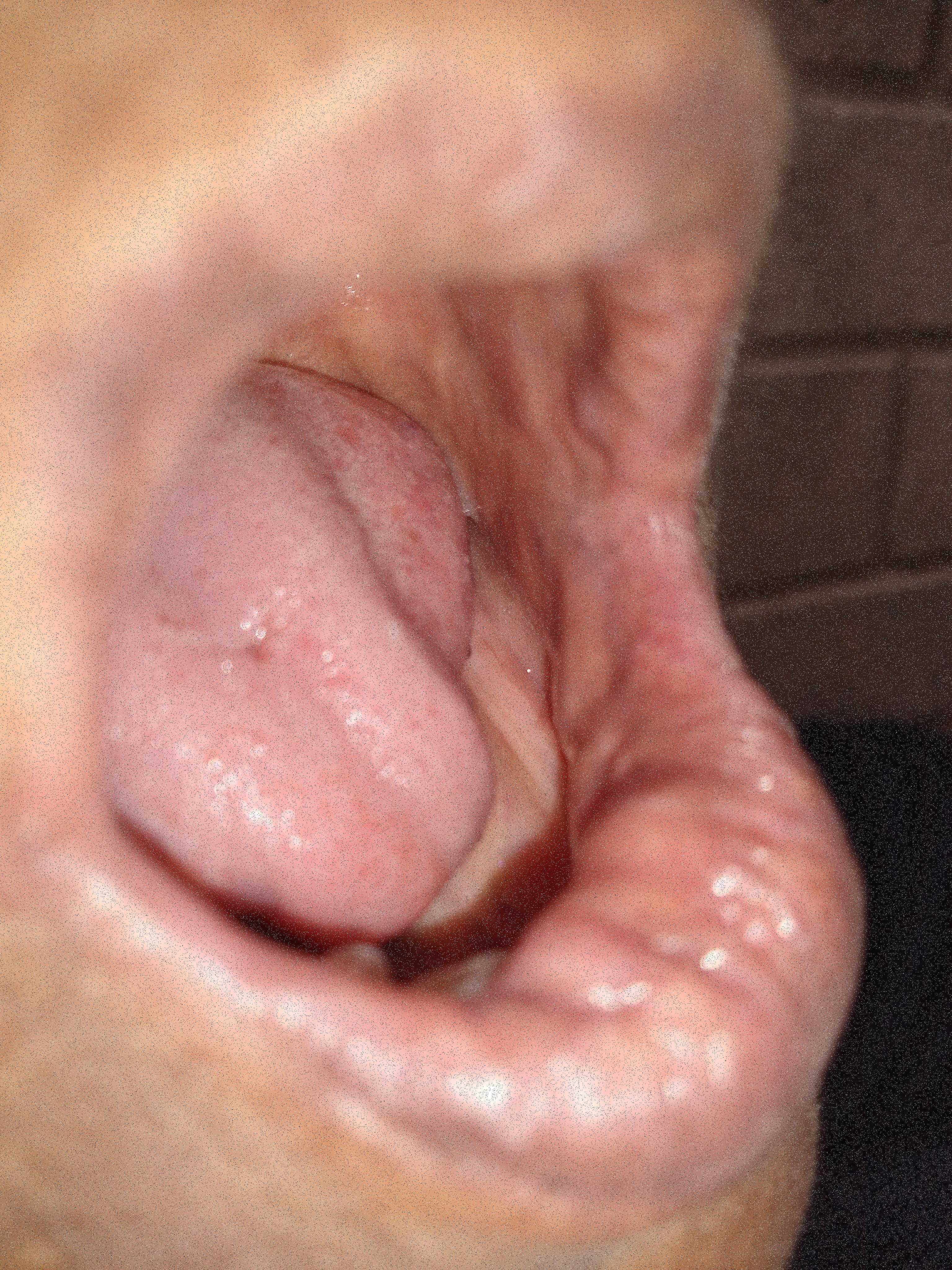}
        \caption{S\&P (5\%)}
        \label{sp}
    \end{subfigure}
    \caption{Original and noisy versions of an image from the test set, with the percentage of synthetic degradation indicated.}
    \label{fig:original_and_noisy_images}
\end{figure}

These perturbations were synthetically introduced at varying intensities to the raw images of the held-out test set. Figure \ref{fig:original_and_noisy_images} illustrates an original sample alongside its degraded counterparts.

\begin{table}[htbp]
\caption{Robustness performance of the optimised model against varying intensities of hardware-simulated noise}
\label{tab: robustness_results}
\centering
\resizebox{\linewidth}{!}{%
\begin{tabular}{|l|c|c|c|}
\hline
\textbf{Noise Type} & \textbf{Intensity (\%)} & \textbf{Sensitivity (\%)} & \textbf{Specificity (\%)} \\ \hline
Baseline (No noise) & 0 & 87.4 & 86.5 \\ \hline
\multirow{3}{*}{Gaussian} 
 & 15 & 76.0 & 83.4 \\ 
 & 20 & 68.8 & 86.4 \\
 & 25 & 58.4 & 89.5 \\ \hline
\multirow{3}{*}{Stripe} 
 & 2 & 80.8 & 82.4 \\ 
 & 5 & 64.6 & 85.8 \\
 & 7 & 51.7 & 88.3 \\ \hline
\multirow{3}{*}{Salt and Pepper} 
 & 2 & 68.5 & 88.1 \\ 
 & 5 & 25.9 & 96.9 \\
 & 7 & 5.1 & 99.3 \\ \hline
\end{tabular}%
}
\end{table}

As detailed in Table \ref{tab: robustness_results}, the system demonstrated a predictable, monotonic degradation under unstructured Gaussian noise, maintaining over 68\% sensitivity even at a severe 20\% noise intensity. This indicates that the hybrid architecture is robust to distributed, low-level sensor noise. 

The model's sensitivity to structured (stripe) noise was non-linear and acute, as expected. The model performance at low-intensity (2\%) stripe noise was comparable to 15\% Gaussian noise, and further increasing the intensity to 7\% caused a significant drop in sensitivity. This can be attributed to the structure of stripe noise, which can mask lesion areas, potentially resulting in missed detections.

The model also exhibited vulnerability to impulse (salt-and-pepper) noise. At 7\% intensity, sensitivity collapsed to 5.1\%, while specificity inflated to 99.3\%. Salt-and-pepper noise introduces high-frequency local disturbances throughout the image, which are spatially propagated during image resizing. This process obliterates distinctive structural features, such as edges and textures, in the preprocessed images passed to the model. Consequently, diagnostic features associated with suspicious lesions were obscured, which led the model to default to a non-suspicious inference, thus resulting in collapsed sensitivity and artificially saturated specificity.
 
Overall, these stress tests confirm the model's operational viability against standard low-light sensor noise (Gaussian), but exhibit explainable performance degradation against structured (stripe) and impulse (salt-and-pepper) noise.


\section{Discussion}
\label{discussion}

This study aimed to develop, optimise, and stress-test lightweight deep learning systems for the binary classification of oral cavity lesions, specifically aimed at resource-constrained, point-of-care edge deployments. While the high variance in smartphone camera specifications and unconstrained lighting conditions posed significant data-curation challenges, relying on this heterogeneous dataset enhanced the generalisability of the final system, forcing it to learn robust features rather than device-specific artefacts.

\subsection{Architectural Findings and Pipeline Efficiency}
Our evaluations highlight that hybrid architectures, specifically MobileViTv2, offer the optimal balance of computational efficiency and diagnostic performance. The success of this model is largely attributable to its separable self-attention mechanism, which computes global context with a linear complexity of $O(N)$ rather than the quadratic $O(N^2)$ bottleneck of standard Vision Transformers. This allows the system to achieve high-resolution spatial awareness while adhering to the stringent memory and compute limits of smartphone inference. 

Furthermore, our pipeline ablation studies revealed that off-the-shelf foundation models, such as the Segment Anything Model (SAM), are ill-suited for segmenting oral cavity images in a zero-shot setting, as the generated masks frequently occluded relevant mucosal boundaries, introducing noise rather than clarity.

Similarly, our exploration of Knowledge Distillation (KD) demonstrated that transferring representations from high-capacity teacher models (e.g., ViT, EfficientNetV2-L) introduces severe training overhead without yielding commensurate performance gains. The catastrophic model collapse observed during complete distillation indicates that ultra-lightweight edge models lack the parameter space to reliably converge on the complex soft-label distributions generated by massive networks. Consequently, our findings strongly advocate for a streamlined training paradigm for edge-triage systems: directly optimising the lightweight architecture using ground-truth labels rather than forcing it to mimic bloated, compute-intensive architectures.

Future systems-level work could consider eschewing heavy, generalised foundation models in favour of training lightweight architectures on domain-specific tasks. Similarly, our interpretability studies (Section \ref{interpretability}) successfully validated that the hybrid model avoids spurious correlations; however, it also underscores the future necessity of incorporating spatial-attention constraints during the training loop to further penalise background-feature reliance.

\subsection{Potential for Clinical Efficacy at the Edge}
The lightweight model, with potential edge-deployment capabilities, has immense potential to democratise oral cancer screening. Existing literature indicates that primary healthcare workers operating in field settings typically achieve lower diagnostic sensitivities of 43.6\%-60\% with slightly superior specificities of 78\%-81\% \cite{birur2022field}. In contrast, our optimised DL system achieves a specialist-benchmarked sensitivity of 87.4\% and a specificity of 86.5\% on the held-out test image dataset, with on-device inference capabilities. 

Critically for a triage system, the best-performing model checkpoint achieved a Negative Predictive Value (NPV) of 97.2\%. In resource-constrained healthcare networks, false positives strain secondary care facilities, but false negatives pose a fatal risk by delaying critical referrals to secondary care. The high NPV indicates that the edge system effectively rules out non-suspicious cases, ensuring that minimal true positives are missed. Furthermore, the trained models are 17 MB in size, enabling easy integration into any mobile application, even on a basic smartphone. Such a lightweight model would require minimal CPU usage, i.e. a peak usage of 1-2 cores for a few milliseconds, drain a negligible amount of battery, and produce inference within a second. A solution capable of offline performance on edge devices, such as smartphones, holds massive promise for providing real-time, specialist-level assistance to frontline health workers in resource-constrained settings.

\subsection{Limitations and Future Directions}

The dataset used for model development is derived from a single country, India. However, the screening programmes through which the data were sourced were conducted over a decade, across a wide geographic area. Moreover, India is a diverse country, and a robust solution for India will likely be robust for other countries with similar issues. Further, due to data anonymisation, there was no patient-level association for the images, i.e., we did not know which images came from the same individual. Therefore, the models could not learn patient-level features, and they relied strictly on RGB pixel data. Future iterations of this work could consider multimodal fusion architectures capable of processing textual patient metadata (e.g., age, sex, relevant habit history) alongside image tensors to provide more comprehensive inference. As a next step, a more granular multi-class classification framework that emphasises learning finer features could further improve the detection of high-risk lesions and optimise resource allocation.

It should be noted that no inter-annotator agreement could be reported because the dataset was collected in program mode and only one specialist reviewed each image. This could lead to subjectivity in interpretations and introduce biases. Also, specialist-segmented lesion masks were available for only a small number of images (n=244). As the interpretability mechanism was based on GradCAM++ and attention rollout, a quantitative interpretability performance measure could not be derived from this small, specialist-annotated subset; thus, a qualitative interpretability analysis was performed. Furthermore, the current models have not been validated on external datasets. Future work could consider external validation with multiple annotators and assess annotator agreement for both classification and lesion segmentation to obtain a stronger ground truth. 

The robustness study in section \ref{robustness} revealed vulnerabilities in the model to structured (stripe) and unstructured (salt-and-pepper) noise. Techniques such as frequency filtering via wavelet decomposition or directional $L_1$-norm optimisation of gradient spikes for destriping, and median filtering or its variants for denoising, can be explored to address performance drops in the presence of structured and unstructured noise in images, thereby helping prevent predictable failures in the model.

Finally, to address the inevitable drift in hardware sensors over time, future research must evaluate the feasibility of continual learning frameworks. These paradigms would allow models to adapt to shifting demographic data and new smartphone camera signal processors, thereby ensuring long-term systemic robustness and broader clinical utility.


\section{Conclusion}
\label{conclusion}

This study presents the development and rigorous systems-level validation of a lightweight deep learning framework for the point-of-care triaging of oral cavity lesions in resource-constrained environments. The training pipeline ablation studies established that directly optimising lightweight architectures on ground-truth labels achieves performance comparable to large models, but with significantly reduced compute and memory requirements. Furthermore, interpretability analysis via gradient-weighted class activation mapping and multi-scale attention rollout validated that the system largely anchors its inference in clinical features rather than spurious background correlations. Stress testing via synthetic hardware noise injection highlighted robustness against unstructured low-light sensor noise.

Clinically, the optimised edge system achieved a specialist-benchmarked sensitivity of 87.4\% and a Negative Predictive Value of 97.2\%, significantly outperforming baselines of primary healthcare workers. Although future work must address risk stratification and comprehensive multimodal analysis at the patient level for better outcomes, the current framework establishes a robust foundation for future edge deployments. Ultimately, the proposed system has immense potential to enable AI-assisted, edge-native oral cancer screening in resource-limited settings, thereby democratising access to improved screening for under-served populations.

\section*{Acknowledgments}
The authors thank Prathosh A. P., Abhijnya Bhat, Soma Biswas, Leena Chavaan, Sundeep Chepuri, Keerthi Gurushanth, Vishal Halder, Manepalli Naveen Sai Kiran, Prabhdeep Kaur, Lakshmi Krishnan, Pramila Mendonca, Deepika Mishra, Naveen Paluru, Tarun Kumar Sahu, Chandra Sekhar Seelamantula, Rajiv Soundarajan, Sumsum Sunny, and Phaneendra K. Yalavarthy for their valuable guidance, discussions, clinical insights, and support throughout the course of this work. Further, the authors thank the reviewers for their valuable feedback.

Funding acknowledgements: This work was supported by the Ministry of Education, Government of India, through the TANUH-AI Centre of Excellence in Healthcare, the Cisco-IISc Centre for Networked Intelligence, Kotak-IISc AI-ML Centre, and the Biocon Foundation. Biocon Foundation kindly shared the data that enabled this study.

\section*{Data and Code Availability}
Data availability for research purposes is subject to approval from ethics committees. Code is available at \mbox{\url{https://github.com/OCS-Tanuh/Conference-AIMLSystems2026}}.

\bibliographystyle{IEEEtran}
\bibliography{ref.bib} 

\end{document}